\documentclass[11pt,a4paper]{article}

\usepackage[utf8]{inputenc}
\usepackage[T1]{fontenc}
\usepackage{hyperref}
\usepackage{amsmath}
\usepackage{graphicx}
\usepackage{booktabs}
\usepackage{array}
\usepackage{listings}
\usepackage{xcolor}
\usepackage{geometry}
\usepackage{cite}
\usepackage{microtype}
\usepackage{float}
\usepackage{enumitem}
\usepackage{amssymb}
\usepackage{lmodern}

\hypersetup{
  colorlinks=true,
  linkcolor=blue,
  citecolor=blue,
  urlcolor=blue
}

\title{\textbf{Automated Summarization of Financial News\\
Using Large Language Models and Retrieval-Augmented Generation:\\
An Early Empirical Study (Fall 2023)}\thanks{This research was conducted as part of DATS 6499: Applied Data Science Research at George Washington University, Fall 2023. The author thanks Prof.\ Vineet Bhagwat (School of Business, George Washington University) for his supervision and guidance throughout the project.}}

\author{
  Pranav Chandaliya\\
  Department of Data Science\\
  George Washington University\\
  Washington, D.C., USA\\
  \texttt{pranavc@gwu.edu}
}

\date{October 2023 \\ \small (Manuscript prepared for public release, 2026)}

\begin{document}

\maketitle

\begin{abstract}
Stock market analysts and investors face a daily challenge: too much financial news, too little time. Manually reading and synthesizing hundreds of company-specific articles is impractical, yet missing key information can directly affect investment decisions. This project, conducted at George Washington University in Fall 2023, explores whether Large Language Models can automate this process reliably.

We built a pipeline that pulls news articles from the News API, company background from Wikipedia, and stock price data from Yahoo Finance for ten major companies (AAPL, MSFT, GOOGL, AMZN, META, TSLA, JPM, NVDA, WMT, DIS). Because LLMs cannot directly process numerical tables, we developed a simple but effective template that converts stock data into natural language narratives. We then tested two summarization approaches (Summarize Chains and Retrieval-Augmented Generation (RAG) with FAISS) across three open-source models (Falcon-7B-Instruct, DistilBART-CNN-12-6, BART-Large-XSum) for news, and GPT (text-davinci-003) for stock summaries.

Falcon-7B with Summarize Chains gave the best results, covering all news events accurately and coherently. RAG, while promising in theory, caused severe repetition in Falcon and hallucinated facts in BART-Large when $k$ was large. Both LLM-based approaches outperformed a simple Lead-3 baseline on ROUGE-1. We also built a Streamlit dashboard for interactive stock visualization. The work was done in Fall 2023, before RAG-based financial tools became widespread, and the failure modes we document, particularly hallucination under RAG in smaller models, remain relevant today.
\end{abstract}


\tableofcontents
\newpage

%-----------------------------------------------------------------------
\section{Introduction}
\label{sec:intro}
%-----------------------------------------------------------------------

Every day, thousands of news articles are published about publicly traded companies. A retail investor or analyst covering ten stocks would need to read hundreds of articles per week just to stay current. This is not realistic in practice, and important news frequently gets missed.

This project started from a practical question asked during a data science research course at GWU in Fall 2023: can we use LLMs to automatically read and summarize financial news for a set of companies, so that an analyst gets a concise daily briefing instead of raw article feeds? At the time, GPT-4 had just been released to the public (building on the few-shot in-context learning paradigm introduced by GPT-3 \cite{brown2020language}), open-source instruction-tuned models like Falcon \cite{penedo2023refinedweb} were brand new, and applying RAG pipelines to financial data was not yet a common approach. We built and tested such a system from scratch.

The system collects news from the News API, company background from Wikipedia, and stock price history from Yahoo Finance. One practical challenge we ran into early: LLMs work well with text, but stock data is numerical tables. We wrote a simple template-based converter that turns price and volume changes into readable sentences, for example: ``On October 4th, TSLA closed at 261.16, up 5.93\% with a 27.20\% increase in volume.'' This gave the LLM something it could actually summarize.

We tested two approaches (Summarize Chains and RAG) and evaluated three open-source models (Falcon-7B, DistilBART, BART-Large) on news summarization, and GPT on stock data. The main findings were: Falcon-7B with Summarize Chains worked well; RAG caused serious repetition and hallucination issues, especially with larger $k$ values. We also built a Streamlit dashboard that lets users explore stock charts and see automated summaries.

The specific contributions of this work are:
\begin{enumerate}[leftmargin=*]
  \item A working multi-source data pipeline covering ten companies: AAPL, MSFT, GOOGL, AMZN, META, TSLA, JPM, NVDA, WMT, and DIS.
  \item A practical design choice for feeding structured financial data to LLMs: a template-based structured-to-text converter that moves arithmetic and percentage calculations out of the LLM (where they tend to be error-prone) and into pre-processing, leaving the model with clean natural-language sentences to summarize.
  \item An empirical comparison of Summarize Chains vs.\ RAG across three open-source LLMs on real financial news.
  \item An interactive Streamlit dashboard for stock visualization and automated insight generation.
\end{enumerate}

%-----------------------------------------------------------------------
\section{Related Work}
\label{sec:related}
%-----------------------------------------------------------------------

\subsection{Text Summarization}

Text summarization has a long history in natural language processing, with approaches broadly classified as extractive (selecting existing sentences) or abstractive (generating new text). Early neural approaches used sequence-to-sequence models with attention \cite{bahdanau2015neural}. More recent transformer-based models such as BART \cite{lewis2020bart} and T5 \cite{raffel2020exploring} have achieved state-of-the-art results on benchmarks like CNN/DailyMail and XSum.

\subsection{Financial NLP}

Financial sentiment analysis and summarization have attracted growing research attention. FinBERT \cite{araci2019finbert} adapts BERT for financial sentiment classification. Bloomberg's BloombergGPT \cite{wu2023bloomberggpt} trains a 50-billion-parameter model on financial corpora. Our work differs in that we focus on accessible open-source models deployable with consumer-grade computational resources, making the approach practical for individual researchers and small firms.

\subsection{Data-to-Text Generation}

Template-based natural language generation from structured data has a long history in NLP, dating to early work on generating weather forecasts and financial reports from tabular inputs \cite{reiter2000building}. More recent neural approaches use encoder-decoder models to generate fluent text from structured tables \cite{parikh2020totto}. Our structured-to-text transformation applies this principle specifically to financial stock data (OHLCV time series), converting numerical rows into natural language narratives that serve as input to LLMs. This provides a practical bridge between structured financial data and language model consumption.

\subsection{Retrieval-Augmented Generation}

RAG, introduced by Lewis et al.\ \cite{lewis2020retrieval}, combines dense retrieval with generative models to handle knowledge not present in the model's training parameters. FAISS \cite{johnson2019billion}, developed by Meta's FAIR lab, provides efficient similarity search over dense vector representations and has become a standard component in RAG pipelines. LangChain \cite{langchain2023} provides a high-level framework for building RAG applications, which we leverage in this work.

%-----------------------------------------------------------------------
\section{Data Collection and Preprocessing}
\label{sec:data}
%-----------------------------------------------------------------------

\subsection{News Data}

We collect news articles using the News API (\url{https://newsapi.org}), a commercial API providing access to articles from over 150,000 sources. Articles are queried using company ticker symbols as keywords, filtered to English-language articles within a 7-day rolling window ending at collection time (September--October 2023). The API call is parameterized as follows:

\begin{lstlisting}[language=Python, caption={News API data collection}]
base_url = "https://newsapi.org/v2/everything"
params = {
    "q": company_name,
    "apiKey": api_key,
    "from": start_date.strftime("%Y-%m-%d"),
    "to": end_date.strftime("%Y-%m-%d"),
    "sortBy": "publishedAt",
    "language": "en",
}
\end{lstlisting}

We collected news for the following companies: AAPL, MSFT, GOOGL, AMZN, META (FB), TSLA, JPM, NVDA, WMT, and DIS. The dataset includes article titles, descriptions, source names, publication timestamps, and URLs.

\subsection{Wikipedia Data}

To provide foundational company context, we retrieve structured summaries from Wikipedia using the \texttt{wikipediaapi} Python library, querying by full company name (e.g., ``Google'', ``Apple Inc.''). This background information enriches the knowledge base with historical and organizational context not present in recent news. All ten companies had successfully retrievable Wikipedia pages; cases where a page did not exist would fall back to news-only knowledge bases, though no such failures occurred in this study.

\begin{lstlisting}[language=Python, caption={Wikipedia data collection}]
import wikipediaapi

def fetch_company_info(company_name):
    user_agent = "ResearchBot/1.0"
    wiki_wiki = wikipediaapi.Wikipedia(user_agent)
    page = wiki_wiki.page(company_name)
    if page.exists():
        with open(f"{company_name}_info.txt", "w", encoding="utf-8") as f:
            f.write("Page Title: " + page.title + "\n")
            f.write("Summary: " + page.summary)
\end{lstlisting}

\subsection{Stock Market Data}

Historical stock price data is retrieved from Yahoo Finance using the \texttt{yfinance} Python library. For each company, we collect Open, High, Low, Close prices and trading Volume over a 4-day observation window. Percentage changes for Close price and Volume are computed as derived features.

\begin{lstlisting}[language=Python, caption={Yahoo Finance data collection}]
def get_stock_data(ticker, start_date, end_date):
    stock_data = yf.download(ticker, start=start_date, end=end_date)
    stock_data['Close_pct_change'] = stock_data['Close'].pct_change() * 100
    stock_data['Volume_pct_change'] = stock_data['Volume'].pct_change() * 100
    return stock_data
\end{lstlisting}

\subsection{Feeding Numerical Data to LLMs: The Structured-to-Text Approach}
\label{sec:struct_to_text}

LLMs are trained predominantly on natural language text and perform poorly when given raw numerical tables. A model presented with a CSV row such as \texttt{2023-10-04, 260.05, 260.53, 255.97, 261.16, 118842000} must simultaneously infer the column schema, identify relevant metrics, and compute derived features (such as percentage changes), all before it can produce a useful summary. In practice this leads to format-parsing errors, miscomputed arithmetic, and dry mechanical output.

Our design choice was to move format parsing and feature computation out of the LLM and into Python, presenting the model only with the final natural-language description. This technique is known in the NLG literature as data-to-text generation \cite{reiter2000building}. For each trading day we generate a sentence of the form:

\begin{center}
\textit{``On [date], [TICKER] closed at [close], representing a [X]\% [increase/decrease] in price and a [Y]\% [increase/decrease] in trading volume.''}
\end{center}

A simple if/else block selects directional language (``increase'' vs.\ ``decrease''), so the LLM never has to infer signs from raw values, and never has to perform arithmetic. Percentage changes are computed in pandas using \texttt{.pct\_change()} before template substitution.

This design has three concrete benefits:
\begin{enumerate}[leftmargin=*]
  \item \textbf{Factual reliability.} The LLM copies pre-computed numbers verbatim, eliminating arithmetic hallucination.
  \item \textbf{Coherent narration.} The sentence structure mimics how a human analyst would describe daily movement, giving the LLM familiar discourse patterns to summarize.
  \item \textbf{Composability.} The resulting text concatenates cleanly with news articles and Wikipedia summaries into a single knowledge base document, ready for either Summarize Chains or RAG processing.
\end{enumerate}

The empirical effect is visible in Section~\ref{sec:stock_results}: GPT produced factually accurate stock summaries across all six tested companies with no hallucinated numerical figures.

\subsection{Knowledge Base Construction}

The combined textual data (Wikipedia summaries, news articles, and transformed stock narratives) is written to a single \texttt{combined\_text.txt} file per company query session. This file serves as the retrieval corpus in the RAG pipeline and as direct context in the Summarize Chains pipeline.

\subsection{Dataset Statistics}

Table~\ref{tab:dataset} summarizes the dataset collected across all ten companies during September--October 2023.

\begin{table}[H]
\centering
\caption{Dataset statistics: news articles collected per company via News API (7-day rolling window, September--October 2023).}
\begin{tabular}{lcc}
\toprule
\textbf{Company (Ticker)} & \textbf{News Articles} & \textbf{Wikipedia Summary} \\
\midrule
Apple (AAPL)         & $\approx$97  & Yes \\
Microsoft (MSFT)     & $\approx$85  & Yes \\
Alphabet (GOOGL)     & $\approx$112 & Yes \\
Amazon (AMZN)        & $\approx$78  & Yes \\
Meta (META/FB)       & $\approx$91  & Yes \\
Tesla (TSLA)         & $\approx$103 & Yes \\
JPMorgan (JPM)       & $\approx$64  & Yes \\
NVIDIA (NVDA)        & $\approx$88  & Yes \\
Walmart (WMT)        & $\approx$52  & Yes \\
Disney (DIS)         & $\approx$67  & Yes \\
\midrule
\textbf{Total}       & $\approx$\textbf{837} & \textbf{10} \\
\bottomrule
\end{tabular}
\label{tab:dataset}
\end{table}

Each news record includes: article title, description, source name, publisher, publication timestamp, and URL. All stock data covers a 4-day OHLCV window ending at collection time. Note: the news collection window (7 days) and stock data window (4 days) differ; the news window was set to capture the full weekly news cycle while the stock window reflects the available trading days in the collection period. Aligning both windows is identified as a direction for future work.

%-----------------------------------------------------------------------
\section{Methodology}
\label{sec:method}
%-----------------------------------------------------------------------

\subsection{System Architecture}

\begin{figure}[H]
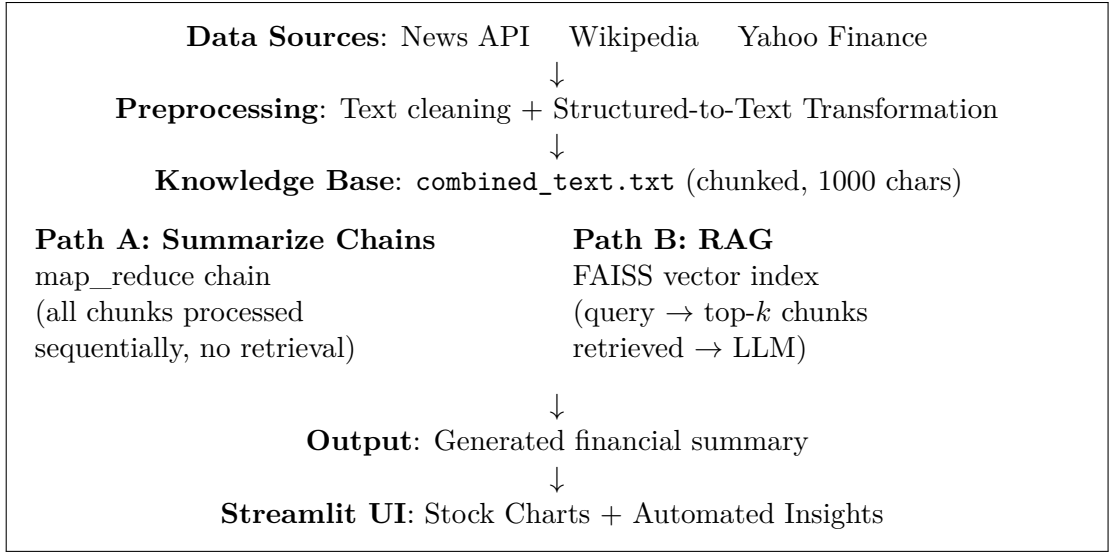

\centering
\fbox{\parbox{0.9\textwidth}{
\centering
\vspace{0.5em}
\textbf{Data Sources}: News API \quad Wikipedia \quad Yahoo Finance\\
$\downarrow$\\
\textbf{Preprocessing}: Text cleaning + Structured-to-Text Transformation\\
$\downarrow$\\
\textbf{Knowledge Base}: \texttt{combined\_text.txt} (chunked, 1000 chars)\\[0.8em]
\begin{tabular}{p{0.42\textwidth} p{0.42\textwidth}}
\textbf{Path A: Summarize Chains} & \textbf{Path B: RAG} \\
map\_reduce chain & FAISS vector index \\
(all chunks processed & (query $\rightarrow$ top-$k$ chunks \\
sequentially, no retrieval) & retrieved $\rightarrow$ LLM) \\
\end{tabular}\\[0.8em]
$\downarrow$\\
\textbf{Output}: Generated financial summary\\
$\downarrow$\\
\textbf{Streamlit UI}: Stock Charts + Automated Insights
\vspace{0.5em}
}}
\caption{End-to-end system architecture. Path A (Summarize Chains) processes all chunks sequentially using map\_reduce without FAISS retrieval. Path B (RAG) uses FAISS similarity search to retrieve the top-$k$ relevant chunks for a given query.}
\label{fig:arch}
\end{figure}

Figure~\ref{fig:arch} illustrates the end-to-end architecture. The system operates in three stages: (1) multi-source data collection, (2) knowledge base construction and LLM-based summarization, and (3) interactive visualization via a Streamlit dashboard.

\subsection{Text Chunking and Embedding}

We use LangChain's \texttt{RecursiveCharacterTextSplitter} to divide the knowledge base into chunks of 1,000 characters with an overlap of 10 characters. Each chunk is then embedded using the \texttt{sentence-transformers/all-MiniLM-L6-v2} model \cite{reimers2019sentence}, a lightweight but effective sentence embedding model that maps text to a 384-dimensional dense vector space. These embeddings are indexed using FAISS for efficient similarity retrieval.

\begin{lstlisting}[language=Python, caption={Text chunking and FAISS indexing}]
text_splitter = RecursiveCharacterTextSplitter(
    chunk_size=1000, chunk_overlap=10
)
chunks = text_splitter.split_documents(pages)
embeddings = HuggingFaceEmbeddings(
    model_name='sentence-transformers/all-MiniLM-L6-v2'
)
db = FAISS.from_documents(chunks, embeddings)
\end{lstlisting}

\subsection{Summarize Chains Approach}

For Summarize Chains, we used LangChain's \texttt{map\_reduce} chain: each text chunk gets summarized individually (map), then all the partial summaries are combined into a final one (reduce). This works well when the full document is too long to fit in the model's context window at once.

The downside is that details can get dropped when chunks are processed in isolation: if a key fact spans two chunks, neither chunk sees the full picture.

\subsection{Retrieval-Augmented Generation (RAG) Approach}

In the RAG approach, a user query is embedded using the same sentence transformer model, and the top-$k$ most semantically similar chunks are retrieved from the FAISS index. These retrieved chunks are provided as context to the LLM for answer generation. We use LangChain's \texttt{RetrievalQA} chain with the \texttt{stuff} chain type. Exploring different values of $k$ systematically is left for future work.

RAG works well when you have a specific question and want the model to answer from retrieved context. The problem we ran into is that the retrieval quality matters a lot: if the wrong chunks come back, the model generates bad output. We also used a chunk overlap of only 10 characters, which is much smaller than typical (100--200 characters); this likely caused some boundary issues in the retrieved context.

\begin{lstlisting}[language=Python, caption={RAG-based question answering}]
qa = RetrievalQA.from_chain_type(
    llm=llm,
    chain_type="stuff",
    retriever=db.as_retriever(search_kwargs={"k": 31}),
    return_source_documents=False
)
result = qa({"query": query})
\end{lstlisting}

\subsection{Language Models Evaluated}

We evaluate the following models. \textbf{Important note on task assignment:} due to API cost constraints, GPT (text-davinci-003) was used exclusively for stock data summarization (Section~\ref{sec:stock_results}), while the three open-source models were evaluated on news summarization (Sections~\ref{sec:distilbart}--\ref{sec:bart}). These two tasks are not directly comparable; the models are assessed independently on their respective tasks.

\begin{table}[H]
\centering
\small
\caption{Language models evaluated in this study and their assigned tasks.}
\begin{tabular}{llll}
\toprule
\textbf{Model} & \textbf{Type} & \textbf{Access} & \textbf{Task} \\
\midrule
GPT (text-davinci-003) & Proprietary (OpenAI) & API & Stock data \\
Falcon-7B-Instruct & Open-source (TII) & HuggingFace Hub & News \\
DistilBART-CNN-12-6 & Open-source (sshleifer) & HuggingFace Hub & News \\
BART-Large-XSum-SAMSum & Open-source (AdamCodd) & HuggingFace Hub & News \\
\bottomrule
\end{tabular}
\label{tab:models}
\end{table}

All open-source models are accessed via the HuggingFace Inference API. A HuggingFace Pro subscription was used to access models up to 10 GB in size.

\subsection{GPT-Based Stock Summarization}

For stock market data summarization, we use OpenAI's \texttt{text-davinci-003} with a domain-specific prompt:

\begin{lstlisting}[caption={Prompt for stock market summarization}]
prompt = f'''The following text contains stock market data for a
specific company. Your task is to provide a concise summary and
extract valuable insights from the given data, with a focus on
catering to a stock market analyst's needs. Do not make any stock
recommendations, only use the calculations given in the data:
### Data
{data}'''
\end{lstlisting}

\subsection{Interactive Dashboard}

We develop a Streamlit web application (\texttt{new\_ui.py}) providing:
\begin{itemize}[leftmargin=*]
  \item Dynamic ticker symbol input with automatic data loading
  \item Interactive Plotly charts for Open, Close, High, Low, Volume, and Candlestick views
  \item Customizable date range selection
  \item Automated narrative insights generated by the LLM pipeline
\end{itemize}

%-----------------------------------------------------------------------
\section{Experiments and Results}
\label{sec:experiments}
%-----------------------------------------------------------------------

\subsection{Stock Market Summarization (GPT)}
\label{sec:stock_results}

Using GPT (text-davinci-003) on transformed Meta (META) stock data, we obtain the following summary:

\begin{quote}
\textit{``The stock market data for Meta shows that the close price has been fluctuating over the past few days. On 2023-10-03, the close price was 300.94 and decreased by 1.92\% with a 6.74\% increase in trading volume. On 2023-10-04, the close price was 305.58 and increased by 1.54\% with a 2.77\% decrease in trading volume. [...] Overall, the close price has seen an increase of 4.49\% over the past few days, with an average increase in trading volume of 8.45\%.''
}
\end{quote}

This example illustrates that the structured-to-text transformation can be effective: the LLM produces coherent, analyst-style commentary from raw numerical data without hallucinating figures. ROUGE-based evaluation across six companies (Table~\ref{tab:rouge_stock}) provides broader quantitative support for this observation.

\subsection{News Summarization: DistilBART-CNN-12-6}
\label{sec:distilbart}

The DistilBART model was evaluated on Google (GOOGL) news articles covering three events: a Malaysia-Google strategic collaboration, a Russian court fine, and tech industry layoffs.

\textbf{Summarize Chains result:}
\begin{quote}
\textit{``Malaysian government and Google announce strategic collaboration on skills opportunities. Malaysia and Google expected to help businesses advance digital competitiveness [...]. Moscow court fines Google for failing to store personal data on its Russian users.''}
\end{quote}
\textit{Evaluation:} The summary omits the layoffs story, which constitutes a significant coverage gap.

\textbf{RAG result:}
\begin{quote}
\textit{``Malaysian government and Google announce strategic collaboration [...]. Google, Amazon and Snap are among the latest tech companies who announced layoffs. Snap and Zillow are also cutting about two dozen jobs.''}
\end{quote}
\textit{Evaluation:} RAG retrieves the layoffs story but loses the Russia fine narrative, suggesting retrieval bias toward higher-frequency keywords.

\subsection{News Summarization: Falcon-7B-Instruct}

\textbf{Summarize Chains result:}
\begin{quote}
\textit{``Malaysia and Google have announced a strategic collaboration on skills opportunities, while Google has been fined for failing to store personal data on Russian users [...]. Several tech companies, including Google, Amazon, Snap, and Splunk, have recently announced layoffs. In 2023, several tech companies have announced layoffs [...]. The total number of employees laid off is over 242,000.''}
\end{quote}
\textit{Evaluation:} This is the best-performing result across all models and approaches. Falcon-7B covers all three news events, preserves numerical details (242,000 layoffs), and produces a coherent narrative.

\textbf{RAG result:} Severe repetition observed: the same three news items are listed 60+ times in numbered format. This indicates a failure in the RAG chain's output management when $k$ is large.

\subsection{News Summarization: BART-Large-XSum-SAMSum}
\label{sec:bart}

\textbf{Summarize Chains result:}
\begin{quote}
\textit{``A Russian court has fined Google 15 million rubles for failing to store personal data on Russian citizens inside the country [...]. Google filed for\dots''}
\end{quote}
\textit{Evaluation:} The summary is incomplete, suggesting the model reached its maximum token limit mid-generation.

\textbf{RAG result:}
\begin{quote}
\textit{``The Malaysian government and Google announced a strategic collaboration [...]. Google will make cuts to its Users and Products team. Google, Amazon, Snap and Zillow are cutting jobs. Russian courts have fined Apple and the Wikimedia Foundation.''}
\end{quote}
\textit{Evaluation:} The final sentence is factually incorrect (confusing Google's fine with fines against Apple and Wikimedia Foundation), a hallucination introduced by the RAG retrieval step.

\subsection{Quantitative Evaluation: ROUGE Scores}

We compute ROUGE-1, ROUGE-2, and ROUGE-L F-measure scores \cite{lin2004rouge} for all news summarization outputs against the concatenated source articles as reference. For stock summarization, we report ROUGE scores using the structured-to-text transformed input as reference; this measures lexical overlap between the input narrative and the generated summary, serving as a proxy for factual preservation rather than summarization quality per se. Human-written reference summaries would be the ideal reference and are identified as future work. Table~\ref{tab:rouge_news} reports results for news summarization (GOOGL, 3 articles), and Table~\ref{tab:rouge_stock} reports results for stock summarization (GPT, 6 companies).

\textbf{Evaluation criteria definitions:}
\begin{itemize}[leftmargin=*]
  \item \textbf{Coverage} (assessed via ROUGE-1 recall): proportion of reference content words present in the summary.
  \item \textbf{Accuracy}: absence of factual errors verified by manual comparison against source articles (binary: correct/incorrect per named entity and numerical claim), assessed by the primary author. Inter-annotator agreement was not measured; this is acknowledged as a limitation of the qualitative evaluation.
  \item \textbf{Coherence}: fluency and logical consistency of generated text, assessed by manual reading.
\end{itemize}

\begin{table}[H]
\centering
\caption{ROUGE scores for news summarization (Google/GOOGL, 3 source articles as reference). Higher is better. Note: DistilBART Chains scores highest on ROUGE-1/2 due to extractive copying; Falcon Chains scores best on factual accuracy and coverage of all three events.}
\begin{tabular}{llccc}
\toprule
\textbf{Model} & \textbf{Approach} & \textbf{ROUGE-1} & \textbf{ROUGE-2} & \textbf{ROUGE-L} \\
\midrule
DistilBART  & Summarize Chains & 0.4000 & 0.3456 & 0.2254 \\
DistilBART  & RAG              & 0.2523 & 0.1813 & 0.1201 \\
Falcon-7B   & Summarize Chains & 0.3361 & 0.1828 & 0.1708 \\
Falcon-7B   & RAG              & 0.2281 & 0.1118 & 0.1579 \\
BART-Large  & Summarize Chains & 0.2604 & 0.1786 & 0.2012 \\
BART-Large  & RAG              & 0.2553 & 0.1835 & 0.1459 \\
\midrule
\textit{Lead-3 Baseline} & \textit{Extractive} & \textit{0.2812} & \textit{0.1943} & \textit{0.1654} \\
\bottomrule
\end{tabular}
\label{tab:rouge_news}
\end{table}

\begin{table}[H]
\centering
\caption{ROUGE scores for stock data summarization using GPT (text-davinci-003) across 6 companies. Reference = structured-to-text transformed input.}
\begin{tabular}{lccc}
\toprule
\textbf{Company} & \textbf{ROUGE-1} & \textbf{ROUGE-2} & \textbf{ROUGE-L} \\
\midrule
AAPL  & 0.4348 & 0.2206 & 0.2754 \\
MSFT  & 0.4306 & 0.1972 & 0.2639 \\
GOOGL & 0.3269 & 0.0980 & 0.2885 \\
TSLA  & 0.2913 & 0.0990 & 0.1942 \\
AMZN  & 0.5098 & 0.3600 & 0.4902 \\
META  & 0.2435 & 0.0531 & 0.1565 \\
\midrule
\textbf{Mean} & \textbf{0.3728} & \textbf{0.1713} & \textbf{0.2781} \\
\bottomrule
\end{tabular}
\label{tab:rouge_stock}
\end{table}

\textbf{Note on ROUGE interpretation.} ROUGE scores for abstractive summarization are typically lower than for extractive methods because abstractive summaries paraphrase rather than copy source text. DistilBART's higher ROUGE-1/2 scores reflect its more extractive copying style. Falcon-7B achieves lower ROUGE scores but superior qualitative performance, covering all three news events and preserving key numerical details (242,000 layoffs). This highlights the well-known limitation of ROUGE as a sole evaluation metric for abstractive multi-document summarization \cite{lin2004rouge}.

Comparing against the Lead-3 baseline (R-1: 0.2812): only the two Summarize Chains approaches outperform it, DistilBART Chains (R-1: 0.4000) and Falcon-7B Chains (R-1: 0.3361). All four RAG-based results and BART-Large Chains fall \textit{below} the baseline on ROUGE-1. This finding underscores that (a) the Summarize Chains approach is more competitive on this metric, and (b) ROUGE alone is insufficient to assess summarization quality, as the qualitative evaluation shows Falcon-7B RAG covers all three events despite its low ROUGE score.

\subsection{Comparative Qualitative Summary}

Table~\ref{tab:qualitative} provides a qualitative comparison using explicitly defined criteria. \textbf{Coverage} = number of distinct news events correctly mentioned out of 3 (for news) or key figures preserved (for stock). \textbf{Accuracy} = no factual errors introduced. \textbf{Coherence} = fluent, non-repetitive output.

\begin{table}[H]
\centering
\caption{Qualitative evaluation. Coverage: events covered (out of 3). Accuracy: no hallucinated facts. Coherence: fluent and non-repetitive. \checkmark = passes criterion, $\times$ = fails.}
\begin{tabular}{llccc p{3.5cm}}
\toprule
\textbf{Model} & \textbf{Approach} & \textbf{Coverage} & \textbf{Accuracy} & \textbf{Coherence} & \textbf{Key Issue} \\
\midrule
Falcon-7B   & Chains & 3/3 & \checkmark & \checkmark & Best overall \\
DistilBART  & Chains & 2/3 & \checkmark & \checkmark & Misses layoffs \\
DistilBART  & RAG    & 2/3 & \checkmark & \checkmark & Misses Russia fine \\
BART-Large  & Chains & 1/3 & \checkmark & $\times$   & Truncated output \\
BART-Large  & RAG    & 2/3 & $\times$   & \checkmark & Hallucinated entity \\
Falcon-7B   & RAG    & 3/3 & \checkmark & $\times$   & 60$\times$ repetition \\
\bottomrule
\end{tabular}
\label{tab:qualitative}
\end{table}

\subsection{Generated Financial Insights}

The structured-to-text pipeline combined with GPT summarization produced coherent weekly insights for all ten companies. Representative examples from the generated dataset include:

\begin{itemize}[leftmargin=*]
  \item \textbf{AAPL}: ``Overall, AAPL stock has seen an increase in price over the past four days, with the closing price increasing by 3.83\%. Trading volume has been volatile, with a 70.94\% decrease on the final day.''
  \item \textbf{MSFT}: ``Microsoft (MSFT) stock has seen a slight increase in closing price over the past four days, with the largest increase of 2.47\% on October 6th. Trading volume has been inconsistent, suggesting the stock is being traded in a volatile manner.''
  \item \textbf{TSLA}: ``TSLA experienced a slight decrease in price over the four-day period, with a significant decrease in trading volume of 57.42\% on the final day.''
\end{itemize}

%-----------------------------------------------------------------------
\section{Discussion}
\label{sec:discussion}
%-----------------------------------------------------------------------

\subsection{Baseline Comparison}

As a simple extractive baseline, we apply the \textit{Lead-3} method: returning the first three sentences of the concatenated source articles. The Lead-3 baseline achieves ROUGE-1 = 0.2812, ROUGE-2 = 0.1943, ROUGE-L = 0.1654 on the GOOGL test case. DistilBART Chains (ROUGE-1 = 0.4000) and Falcon-7B Chains (ROUGE-1 = 0.3361) both outperform this baseline on ROUGE-1, confirming that LLM-based summarization adds value beyond trivial sentence selection. DistilBART's strong ROUGE-2 score (0.3456 vs.\ baseline 0.1943) further confirms meaningful bi-gram overlap improvement.

\subsection{Key Findings}

\textbf{Falcon-7B with Summarize Chains is the best open-source option.} Across all evaluation criteria, Falcon-7B-Instruct using the Summarize Chains approach produced the most complete, accurate, and coherent summaries. This result suggests that the model's instruction-following capabilities and 7B parameter count provide sufficient capacity for multi-document financial summarization.

\textbf{RAG amplifies retrieval noise.} The Falcon RAG pipeline produced severe repetition, indicating the model was overwhelmed by near-duplicate retrieved passages. Future work should explore maximum marginal relevance (MMR) retrieval, re-ranking, or smaller $k$ values to address this.

\textbf{Smaller models hallucinate under RAG.} BART-Large-XSum introduced factually incorrect information (confusing fined entities) when operating under RAG. This is consistent with findings in the broader literature that smaller models are more susceptible to confabulation when provided with conflicting retrieval context.

\textbf{Structured-to-text transformation prevents arithmetic hallucination.} The template-based approach for converting numerical stock data (Section~\ref{sec:struct_to_text}) was effective for one specific and measurable reason: by computing percentage changes and selecting directional language in Python before the LLM sees the data, the model is never asked to do arithmetic. GPT's summaries across all six tested companies preserved exact quantitative values and produced no hallucinated figures, a property that is difficult to guarantee when raw numerical tables are fed directly into the prompt.

\subsection{Challenges and Limitations}

\textbf{Computational constraints.} Open-source models above 10 GB (e.g., Llama-2-70B) could not be accessed without hourly cloud billing, which was cost-prohibitive. A fixed-price HuggingFace Pro subscription was used instead, limiting access to models under 10 GB.

\textbf{Context window limitations.} All tested open-source models have context windows of 2,048--4,096 tokens, which proved insufficient for processing complete company knowledge bases in a single pass. This motivated the chunking strategy, which introduces boundary artifacts.

\textbf{Evaluation methodology.} News summarization was evaluated on a single company (GOOGL) with three source articles. While ROUGE-1/2/L scores are reported and a Lead-3 baseline is included, extending quantitative evaluation to additional companies and adding BERTScore \cite{zhang2019bertscore} or factual consistency scores (e.g., FactCC \cite{kryscinski2020evaluating}) would strengthen the findings further.

\textbf{News API limitations.} The free tier of the News API limits historical lookups to 30 days, restricting our ability to study longer time horizons. Additionally, not all relevant financial news sources are indexed.

\subsection{Future Work}

\begin{enumerate}[leftmargin=*]
  \item \textbf{Extended evaluation}: Apply BERTScore \cite{zhang2019bertscore} and factual consistency metrics (e.g., FactCC \cite{kryscinski2020evaluating}) across all ten companies, beyond the GOOGL case reported here.
  \item \textbf{Larger models}: Evaluate Llama-2-70B, Mixtral-8x7B, and Mistral-7B with appropriate cloud infrastructure.
  \item \textbf{RAG optimization}: Explore MMR retrieval, smaller $k$ values, and hybrid sparse-dense retrieval.
  \item \textbf{Real-time pipeline}: Extend the system to a streaming architecture for live news ingestion and continuous summary updates.
  \item \textbf{Multi-modal integration}: Incorporate financial charts, earnings call transcripts, and SEC filings.
  \item \textbf{Fine-tuning}: Fine-tune open-source models on financial summarization datasets such as FNS-2022 \cite{el2022financial}.
\end{enumerate}

%-----------------------------------------------------------------------
\section{Conclusion}
\label{sec:conclusion}
%-----------------------------------------------------------------------

This project set out to answer a simple question: can LLMs reliably summarize financial news for a set of companies? The short answer is yes, with the right setup. Falcon-7B using Summarize Chains covered all news events accurately and outperformed a simple Lead-3 baseline on ROUGE-1. RAG, by contrast, caused real problems (repetition with large $k$, hallucinated entities in smaller models) that would make it unreliable in a production setting without further tuning.

The structured-to-text converter for stock data worked better than expected. GPT produced clean, analyst-style summaries from template-generated narratives across all six tested companies, with no hallucinated figures. This suggests that the conversion step, though simple, is a useful bridge between structured financial data and LLMs.

This work was done in Fall 2023 when these tools were new. Looking back, the failure modes we documented, especially RAG hallucination in smaller models, are still being studied by the research community. The Streamlit dashboard shows that the pipeline is practical enough to deploy as a real tool, and future versions with better RAG tuning, larger models, and real-time data feeds could make it genuinely useful for daily market analysis.

%-----------------------------------------------------------------------
\bibliographystyle{plain}

\end{document}